\documentclass[10pt,letterpaper]{article}

\usepackage{ccn}
\usepackage{pslatex}
\usepackage{apacite}
\usepackage{epstopdf}
\usepackage{graphicx}
\usepackage{amsmath}
\usepackage{amssymb}

\title{Exploring the Imposition of Synaptic Precision Restrictions For Evolutionary Synthesis of Deep Neural Networks}

\author{{\large \bf M.~J. Shafiee (mjshafiee@uwaterloo.ca), F. Li (f27li@uwaterloo.ca), A. Wong (a28wong@uwaterloo.ca)} \\
 Department of Systems Design Engineering\\
University of Waterloo, Waterloo, Canada}

\begin{document}

\maketitle

\vspace{-0.55cm }
\begin{quote}
\end{quote}
\vspace{-0.55cm }
\section{Introduction}

	Deep neural networks~\cite{lecun2015deep} can be considered one of the most successful biology-inspired machine learning approaches, having shown significant improvements in accuracy over other machine learning approaches for a wide range of challenges ranging from image classification and segmentation~\cite{krizhevsky2012imagenet} to speech recognition and gene sequencing. A key contributing factor to these incredible feats has been the significant rise and proliferation of massively parallel computing power, allowing researchers to greatly increase the size and depth of deep neural networks, leading to significant improvements in modeling accuracy.  As such, while much of research in deep neural networks had focused on the design of deeper, larger, and more complex deep neural networks, less attention and focus on exploring the notion of synaptogenesis, which is the formation of synapses between neurons, in deep neural networks.  However, the exploration of synaptogenesis in deep neural networks can have significant benefits for the formation of efficient deep neural network architectures, which is important for resource-starved scenarios where computational complexity and memory is limited and energy requirements are strict.
	
	Motivated to explore synaptogenesis for forming highly efficient deep neural networks,~\citeA{javad2016evonet} took two key inspirations from nature: I) stochastic behaviour exhibited during synaptogenesis, as observed by~\citeA{Hill} for specific functional connectivity in neocortical neural microcircuits of Wistar rats, and II) evolutionary regression of neurological functionality for energy preservation, as observed by~\cite{moran2015energetic} in the eyeless Mexican cavefish, which evolved to lose its vision system over generations due to the high metabolic cost of vision.  Leveraging these two key inspirations, a new evolutionary synthesis strategy was proposed by~\cite{javad2016evonet,shafiee2016evolutionary} to synthesize increasingly efficient yet effective deep neural networks over successive generations in a stochastic manner, resulting in evolved deep neural networks that can operate well in resource-starved scenarios. Differing from neuroevolution approaches~\cite{stanley2002evolving}, it forgoes classical methods like genetic algorithms or evolutionary programming and instead introduces a probabilistic generative modeling strategy.  One factor that has not been explored within such an evolutionary synthesis framework is that of synaptic precision, which can have significant impact on computational complexity.  In particular, an observation that was made by a number of researchers such as~\citeA{Baldassi} is that, consistent with recent biological findings, near-optimal performance can be achieved in deep neural networks with limited synaptic precision~\cite{gupta2015deep,hubara2016quantized}.  Motivated by this observation, in this study, we explore the imposition of synaptic precision restrictions and its impact on the evolutionary synthesis of deep neural networks.

\vspace{-0.2cm }
\section{Main Idea}
\vspace{-0.1 cm}
The evolutionary synthesis strategy was proposed by~\citeA{javad2016evonet}, where progressively more efficient deep neural networks are synthesized over successive generations in a stochastic manner. The genetic encoding of a deep neural network is represented by a synaptic probability model, which can be viewed as the `DNA' of the network and is used to mimic the notion of heredity.  An offspring network is synthesized stochastically based on this synaptic probability model and a set of computational environmental conditions to establish the new generation. The offspring network  is then trained to reach to its modeling capability. This evolution process is repeated over generations until the desired traits are met.

The combination of genetic information from the previous generation and environmental conditions during the stochastic synthesis process encourages the newly synthesized offspring networks to possess diverse yet effective network architectures. For instance, when the imposed environmental conditions are configured to favor more efficient network architectures within this stochastic synthesis process, the formation of synapses in the offspring networks will be inherently sparser while following a random behavior that mimics the stochastic neuro-plasticity behaviour in the brain as observed by~\citeA{Hill}.  As such, the random, yet environmentally-influenced nature of synaptogenesis in the evolved deep neural networks compensates for the reduction in synapses compared to the previous generations.
\begin{figure}[!th]
	\vspace{-0.25 cm}
	\begin{center}
		\includegraphics[width = 8 cm]{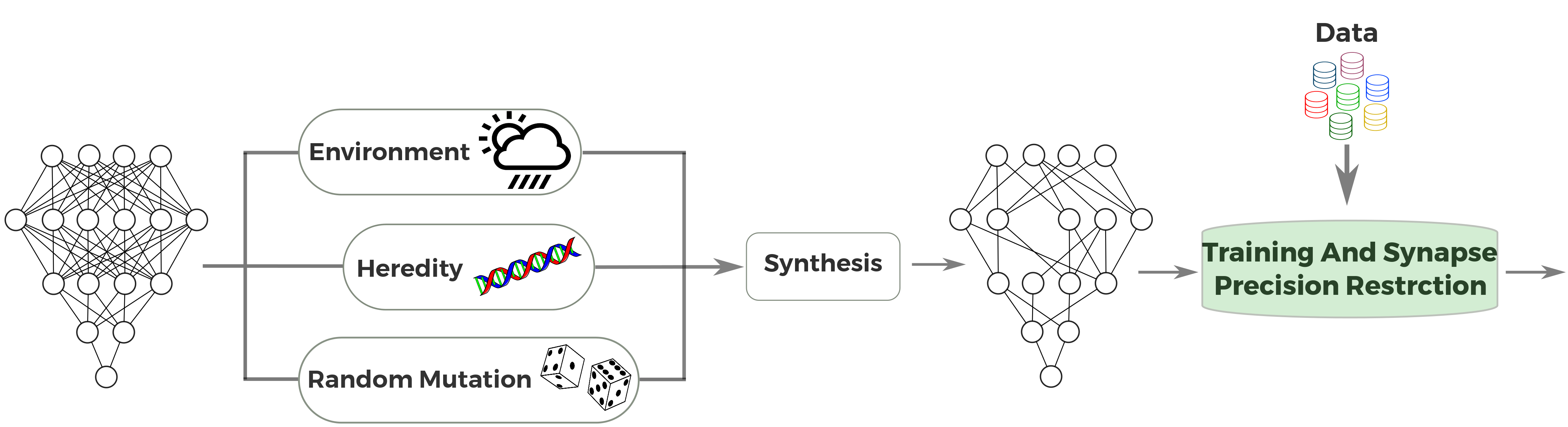}
	\end{center}
	\vspace{-0.45 cm}
	\caption{Overview of evolutionary synthesis of deep neural networks with imposed synaptic precision restrictions: A new offspring network is synthesized stochastically using the `DNA' of the previous generation and environmental factors to mimic heredity, natural selection, and random mutation.  It is then trained, with synaptic precision restriction being imposed.  This process is repeated over generations.}
	\label{fig:flowdiagram}
	\vspace{-0.15 cm}
\end{figure}

Mathematically, the genetic encoding scheme is formulated as $P(S_g|W_{g-1})$, where the network architecture $S_g$ at generation $g$ is influenced by the synaptic strengths $W_{g-1}$ of generation \mbox{$g-1$}.  The random variables $s_i \in S_g$ is a binary random variable with two states $\{0, 1\}$ which specifies whether the synapse $s_i$ exists in the generation $g$ or not. An offspring network is synthesized stochastically via a synthesis probability $P(S_g)$, which combines the synapse probability model $P(S_g|W_{g-1})$ with environmental factors $F(\alpha)$ being imposed:
\begin{align}
P(S_g) \approx P(S_g|W_{g-1}) \cdot F(\alpha).
\end{align}
To explore the notion of imposing synaptic precision restrictions and its impact on the evolutionary synthesis of deep neural networks, we extend the aforementioned process by enforcing a strong synaptic precision constraint at each generation after the training of the offspring network (Figure~\ref{fig:flowdiagram}).  By enforcing this synaptic precision constraint on the trained offspring network, this effectively influences the synaptogenesis behaviour of its offspring deep neural networks.  For this study, a half-precision (16-bit) synaptic precision constraint is enforced at each generation, while training is performed at full precision (32-bits).  The resulting offspring networks have half-precision synaptic precision.  If successful, such impositions of synaptic precision restrictions can result in evolved deep neural networks that further reduce the computational requirements needed for inference.

\vspace{-0.2 cm}
\section{Preliminary Results}
\begin{figure}[!th]
	\vspace{-0.4cm}
	\centering
	\begin{tabular}{ll}
		\includegraphics[width = 4.5 cm]{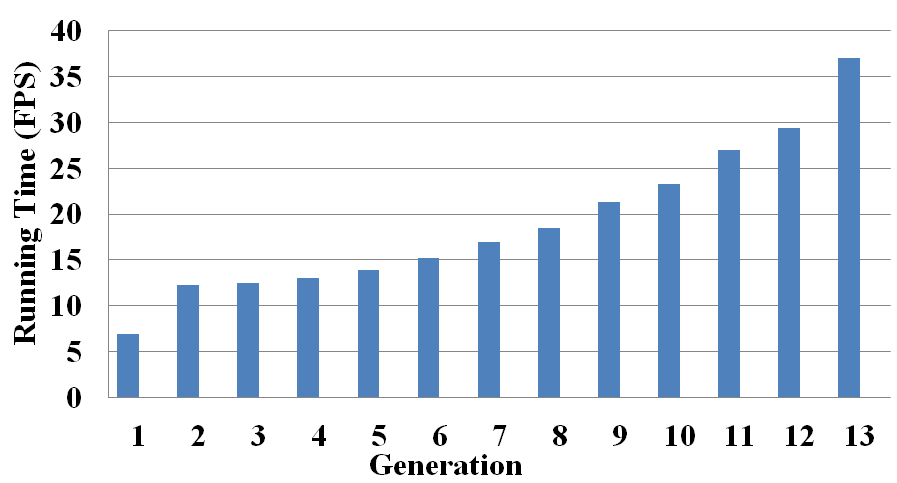}&
		\includegraphics[width = 4.5 cm]{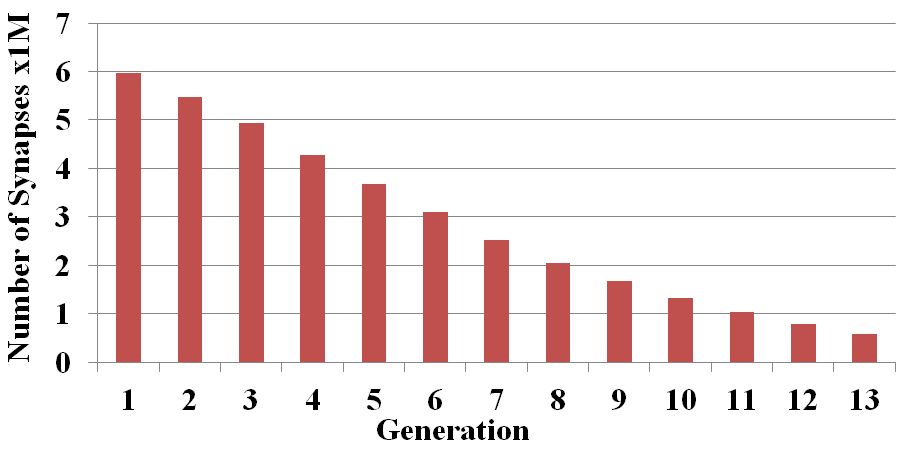}
		\end{tabular}
			\vspace{-0.45 cm}
	\caption{Parse-27k results; As seen the 13th offspring network performs much faster, inferencing at 37 fps compared to the first generation of 7 fps, and with a ten-fold decrease in synapses. }
	\label{fig:carfps}
	\vspace{-0.25 cm}
\end{figure}

\vspace{-0.1 cm}
To study the influence of imposing synaptic precision restrictions on the evolutionary synthesis process, we examine the performance of evolved offspring deep neural networks over generations within the context of object detection using the Parse-27k dataset on the Jetson TX1 within Caffe framework and TensorRT, which supports accelerated half-precision operations. Figure~\ref{fig:carfps} demonstrates the inference speed (in frames per second) and number of synapses for 13 generations of evolved deep neural networks using the evolutionary synthesis process with synaptic precision restriction impositions.  As seen, the number of synapses decreases significantly over generations while the inference speed is increased from 7 fps to 37 fps over generations. As shown in Figure 3, precision and recall is largely consistently between generations (with increase of $\sim$2\% in precision and decrease $\sim$2\% in recall from first generation), which indicates that modeling performance is retained by the offspring deep neural networks.  These results show that the imposition of synaptic precision restrictions within the evolutionary synthesis process can result in evolved deep neural networks with not only significantly reduced number of synapses (a ten-fold decrease in synapses), but also reduced precision requirements while maintaining modeling and inference performance, which can have significant benefits when used in resource-starved environments with limited computational, memory, and energy requirements.

\begin{figure}[!th]
	\centering
	\begin{tabular}{c}
		\includegraphics[width = 6 cm]{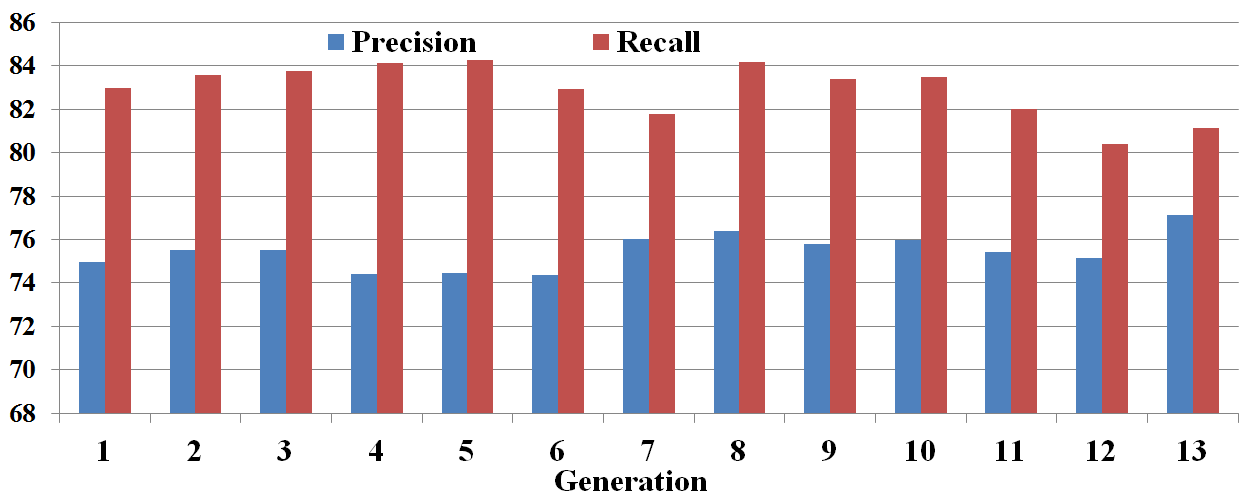}
	\end{tabular}
	\vspace{-0.2 cm}
	\caption{Precision and recall for 13 generations. The precision and recall results demonstrate that modeling performance is retained by the offspring deep neural networks. }
	\label{fig:PeoplePrecision}
	\vspace{-0.5 cm}
\end{figure}
\vspace{-0.35 cm}
\section*{Acknowledgment}
\vspace{-0.15 cm}
The authors thank NSERC, the Canada Research Chairs program, Nvidia, and DarwinAI.
%
\vspace{-0.1 cm}


\bibliographystyle{apacite}

\setlength{\bibleftmargin}{.125in}
\setlength{\bibindent}{-\bibleftmargin}

\bibliography{ccn_style}

\end{document}